\documentclass[preprint]{elsarticle}
\usepackage[a4paper]{geometry}
\usepackage{booktabs}
\usepackage{graphicx}
\usepackage{amsmath}
\usepackage{multirow}
\usepackage{amsfonts}
\usepackage{makecell}
\usepackage{algorithm}
\usepackage{enumitem}
\usepackage{float}
\usepackage{algpseudocode}
\usepackage[toc,page]{appendix}
\usepackage{url}
\usepackage{array}
\usepackage{subcaption}
\usepackage{array,arydshln}
\usepackage{amssymb}
\usepackage{hhline}
\usepackage{times,rotating}
\usepackage{tabularx,ragged2e}
\newcolumntype{T}[1]{>{\raggedright\arraybackslash}p{#1}}
\newcolumntype{M}[1]{>{\centering\arraybackslash}m{#1}}

\newcolumntype{L}[1]{>{\raggedright\let\newline\\\arraybackslash\hspace{0pt}}m{#1}}
\newcolumntype{C}[1]{>{\centering\let\newline\\\arraybackslash\hspace{0pt}}m{#1}}
\newcolumntype{R}[1]{>{\raggedleft\let\newline\\\arraybackslash\hspace{0pt}}m{#1}}

\usepackage{multicol} 

\usepackage{bm} 

\usepackage{rotating}

\usepackage{xcolor}
\definecolor{orange}{HTML}{FFC17D}
\definecolor{blue}{HTML}{7BABFF}
\definecolor{green}{HTML}{A1D68B}

\journal{}

\begin{document}

\begin{frontmatter}

\title{From Data to Actions in Intelligent Transportation Systems: A~Prescription of Functional Requirements for Model~Actionability}

\author[a]{Ibai La\~{n}a\corref{cor1}} 
\author[b]{Javier J. Sanchez-Medina}
\author[c]{Eleni I. Vlahogianni}
\author[a,d]{Javier Del Ser}

\address[a]{TECNALIA, Basque Research \& Technology Alliance (BRTA), P. Tecnologico Bizkaia, Ed. 700, 48160 Derio, Spain}
\address[b]{CICEI, Department of Computer Science, University of Las Palmas de Gran Canaria, 35001 Las Palmas, Spain}
\address[c]{Department of Transportation Planning and Engineering, National Technical University of Athens, 15780 Zografou, Greece}
\address[d]{University of the Basque Country (UPV/EHU), 48013 Bilbao, Spain}
\cortext[cor1]{Corresponding author. TECNALIA, Basque Research \& Technology Alliance (BRTA), P. Tecnologico, Ed. 700. 48170 Derio (Bizkaia), Spain. E-mail: \texttt{ibai.lana@tecnalia.com} (Ibai La\~na).}

\begin{abstract}
Advances in Data Science permeate every field of Transportation Science and Engineering, resulting in developments in the transportation sector that {are} data-driven. Nowadays, Intelligent Transportation Systems (ITS) could be arguably approached as a ``story'' intensively producing and consuming large amounts of data. A~diversity of sensing devices densely spread over the infrastructure, vehicles or the travelers' personal devices act as sources of data flows that are eventually fed {into} software running on automatic devices, actuators or control systems producing, in~turn, complex information flows {among} users, traffic managers, data analysts, traffic modeling scientists, etc. These~information flows provide enormous opportunities to improve model development and decision-making. This work aims to describe how data, coming from diverse ITS sources, can be used to learn and adapt data-driven models for efficiently operating ITS assets, systems and processes; in~other words, for data-based models to fully become \emph{actionable}. Grounded in this described data modeling pipeline for ITS, we~define the characteristics, engineering requisites and challenges intrinsic to its three compounding stages, namely, data fusion, adaptive learning and model evaluation. We~deliberately generalize model learning to be adaptive, since, in~the core of our paper is the firm conviction that most learners will have to adapt to the ever-changing phenomenon scenario underlying the majority of ITS applications. Finally, we~provide a prospect of current research lines within Data Science that can bring notable advances to data-based ITS modeling, which will eventually bridge the gap towards the practicality and actionability of such models.
\end{abstract}


\end{frontmatter}

\section{Introduction}
In the last years Intelligent Transportation Systems (ITS) have experienced an unparalleled expansion for many reasons. The~availability of cost-effective sensor networks, pervasive computation in assorted flavors (distributed/edge/fog computing) and the so-called Internet of Things are all accelerating the evolution of ITS~\cite{zhu2018big}. On top of them, Smart~Cities cannot be understood anyhow without Smart Mobility and ITS as technological pillars sustaining their operation~\cite{albino2015smart}. Smartness springs from connectivity and intelligence, which implies that massive flows of information are acquired, processed, modeled and used to enable faster and informed decisions.  

For the last couple of decades, ITS have grown enough to cross pollinate with previously distant areas such as Machine Learning and its superset in the Artificial Intelligence taxonomy: Data Science. These days Data Science is placed at the methodological core of works ranging from traffic and safety analysis, modeling and simulation, to transit network optimization, autonomous and connected driving and shared mobility. Since the early 90's most ITS systems exclusively relied on traditional statistics, econometric methods, Kalman filters, Bayesian regression, auto-regressive models for time series and Neural Networks, to mention a few~\cite{zhang2011data,karlaftis2011statistical}. What has changed dramatically over the years is the abundance of available data in ITS application scenarios as a result of new forms of sensing (e.g.,~crowd sensing) with unprecedented levels of heterogeneity and velocity. Zhang~et~al.~\cite{zhang2011data} have defined this new form of data-driven ITS as the systems that have vision, multisource, and~learning algorithms driven to optimize its performance and augment its privacy-aware people-centric~character.

The exploitation of this upsurge of data has been enabled by advances in computational structures for data storage, retrieval and analysis, which have rendered it feasible to train and maintain extremely complex data-based models. These baseline technologies have laid a solid substrate for the proliferation of studies dealing with powerful modeling approaches such as Deep Learning or bio-inspired computation~\cite{del2019bioinspired}, which currently protrude in the literature as the \emph{de facto} modeling choice for a myriad of data-intensive~applications. 

However, significant consideration must be placed to the systematic and myopic selection of complex data-based solutions over well-established modeling choices. The~current research mainstream seems to be misleadingly focusing on performance-biased studies, in~a fast-paced race towards incorporating sophisticated data-based models to manifold research area, leaving aside or completely disregarding the operational aspects for the applicability of such models in ITS environment. The~scope of this work is to review existing literature on data-driven modeling and ITS, and~identify the functional elements and specific requirements of engineering solutions, which are the ultimate enablers for data-based models to lead towards efficient means to operate ITS assets, systems and processes; in other words, for data-based models to fully become \emph{actionable}. 
Bearing the above rationale in mind, this~work underscores the need for formulating the requirements to be met by forthcoming research contributions around data-based modeling in ITS systems. To this end, we~focus mainly on system-level on-line operations that hinge on data-based pipelines. However, ITS is a wide research field, encompassing operations held at longer time scales (e.g.,~long-term and mid-term planning) that may not demand some of the functional requirements discussed throughout our work. Furthermore, our discussions target system-level operations rather than user-level or vehicle-level applications, since in the latter the information flow from and to the system is scarce. Nevertheless, some of the described functional requirements for system-level real-time decisions can be extrapolated to other levels and time scales seamlessly.
From this perspective, our ultimate goal is to prescribe -- or at least, set forth -- the main guidelines for the design of models that rely heavily on the collection, analysis and exploitation of data. To this end, we~delve into a series of contributions that are summarized below:
\begin{itemize}[leftmargin=*]
	\item  In the first place, we~identify the gap between the data-driven research reported so far, and~the practical requirements that ITS experts demand in operation. We~capitalize on this gap to define what we herein refer to as \emph{actionable data-based modeling workflow}, which comprises all data processing stages that should be considered by any actionable data-based ITS model. Although diverse data-based modeling workflows can be found in literature with different purposes, most of them count on recognized stages, that are presented in this work from an actionability perspective, i.e.,~what to take into account from the operational point of view when designing the workflow, how~to capture and preprocess data, how to develop a model and how to prescribe its output. These guidelines are proposed and argued within an ITS application context. However, they can be useful for any other discipline in which data-based modeling is performed.
	\item Next, functional requirements to be satisfied by the aforementioned workflow are described and framed in the context of ITS systems and processes, with examples exposing their relevance and consequences if they are not fulfilled.The contributions of this section are twofold: on {the} one hand, we~identify and define the holistically actionable ITS model along with its main features; on the other hand, we~enumerate requirements for each feature to be considered actionable, as well as a review of the latest literature dealing with these features and requisites.
	\item Finally, on a prospective note we elaborate on current research areas of Data Science that should progressively enter the ITS arena to bridge the identified gap to actionability. Once the challenges of modeling and ITS requirements have been stated, we~review emerging research areas in Artificial Intelligence and Data Science that can contribute to the fulfilment of such requirements. We~expect that our reflexive analysis serves as a guiding material for the community to steer efforts towards modeling aspects of more impact for the field than the performance of the model itself.
\end{itemize}

{As a summary, the~contributions of this work consist of identifying the main actionability gaps in the data-based modeling workflow, gathering and describing the fundamental requirements for a system to be actionable, and~considering both the requirements and the usual data-based processing workflow, proposing solutions through the most recent technologies.} These contributions are organized throughout the rest of the paper as follows: Section~\ref{pipeline} delves into the \emph{actionable data-based modeling workflow}, i.e.,~the canonical data processing pipeline that should be considered by a fully actionable ITS system with data-based models in use. Section~\ref{requirements} follows by elaborating on the functional features that an ITS system should comply with so as to be regarded as \emph{actionable}. Once these requirements are listed and argued in detail, Section~\ref{challenges} analyzes research paths under vibrant activity in areas related to Data Science that could bring profitable insights in regards to the actionability of data-based models for the ITS field, such~as explainable AI, the~inference of causality from data, online learning and adaptation to non-stationary data flows. Finally, Section~\ref{conclusion} concludes the paper with summarizing remarks drawn from our prospect.

\section{From Data to Actions: An Actionable Data-based Modeling Workflow}\label{pipeline}

ITS applications with data driven modeling problems underneath range from the characterization of driving behavioral patterns, the~inference of typical routes or traffic flow forecasting, among others. Data driven modeling can be considered to include the family of problems where a computational model or system must be characterized or learned from a set of inputs and their expected outputs~\cite{eiben2003introduction}. In~the context of this definition, actionability complements the data-driven model by prescribing the actions (in the form of rules, optimized variable values or any other representation alike) that build upon the output knowledge enabled by the model.

\begin{figure}[H]%
\includegraphics[width=0.97\columnwidth]{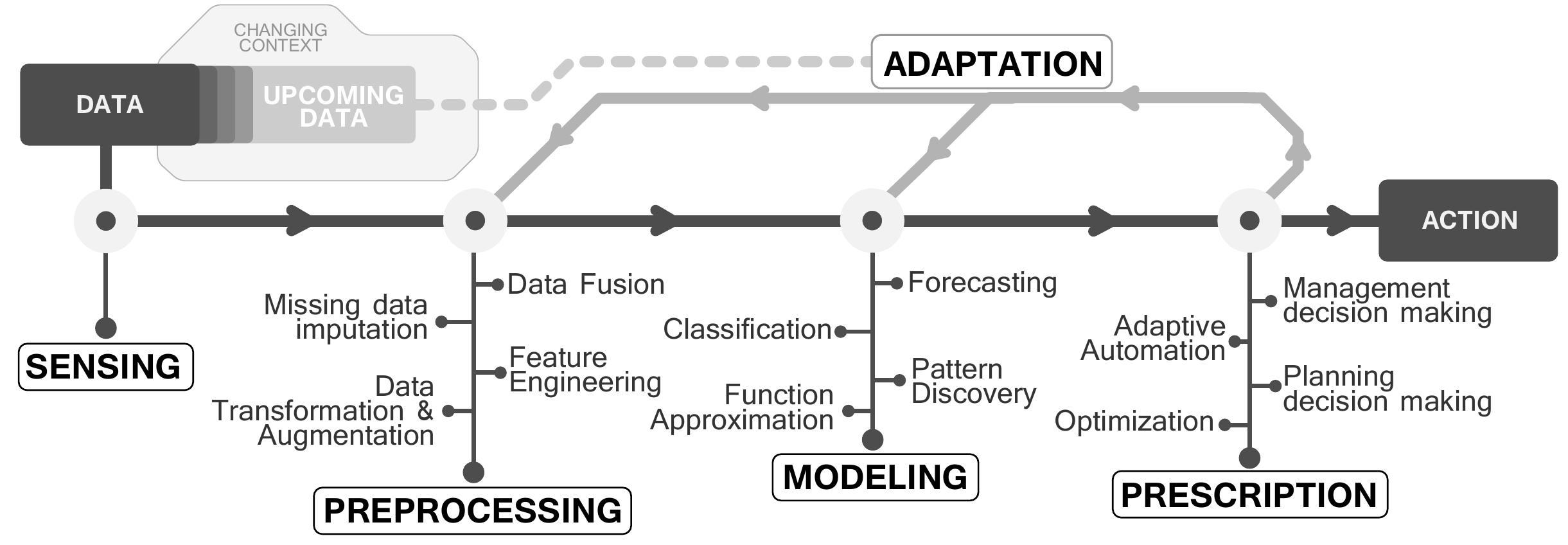}
\caption{Data-based modeling workflow showing its main processing stages and their principal technology~areas.}
\label{fig:1}
\end{figure}

In general, a~design workflow for data-based modeling consists of 4 sequential stages: (1)~data acquisition (\emph{sensing}), which usually considers different sources; (2)~data preprocessing, which aims at building consistent, complete, statistically robust datasets; (3)~data modeling, where a model is learned for different purposes; and (4) model exploitation, which includes the definition of actions to be taken with respect to the insights provided by models in real life application scenarios. These 4 stages can be regarded as the core of off-line data-driven modeling; however, when time dimension joins the game, a~fifth stage -- adaptation -- must be considered as an iterative stage of this data pipeline, aimed at maintaining learned models updated and adapted to eventual changes in the data distribution. This adaptation is crucial for real-life scenarios, where changes can happen in all stages, from~variations of the input data sources, to interpretation adjustments and other sources of non-stationarity imprinting the so-called \emph{concept drift} in the underlying phenomenon to be modelled~\cite{ditzler2015learning}. We~now delve into these five data processing stages in the context of their implementation in ITS applications, following the diagram in Figure \ref{fig:1}.

{ The stages provided in Figure \ref{fig:1} can be considered as a standard workflow in any data-based work; however, although these steps are easily recognisable, they are not always regarded, and~it is common to observe that practitioners put the focus only on a subset of them, disregarding their interactions or omitting some of them. For~instance, the~prescription stage is not frequently considered, while it is an essential link between the modeling outcome and the final decision/action derived from the modeling result. Besides, each step can have implications for the final actionability of the model, reason for which all of them are analyzed below.}

\nointerlineskip

\subsection{Data Acquisition (Sensing)} \label{sensing}
The path towards concrete data-based actions departs from the capture of available ITS information, which in this specific sector is plentiful and highly diverse. The~advent of data science for ITS has come along with the unfolding of copious data sources related to transportation. Indeed, ITS are pouring volumes of sensed data, from~the environment perception layer of intelligent and connected vehicles, to human behaviour detection/estimation (drivers, passengers, pedestrians) and the multiple technologies deployed to sense traffic flow and behaviour. Concurrently, many other non-traditional sources that were useful to infer behavioral needs and expectations of people that use transportation, such~as social media, have started to become increasingly available and exploited augmenting the more conventional sensing sources towards more efficient mobility solutions. Some of these data sources are currently used in almost any domain of ITS, from~operational perspectives such as the estimation of future transportation demands, adaptive signaling or the discovery of mobility patterns, to the provision and of practical solutions, such~as the development of autonomous vehicles{, although not all sources are suitable for all applications. The~model actionability is dependant on this early stage too, reason for which the data selection (when possible) should not be neglected. For~instance, a~model that consumes speed data will probably require some other measurements (maximum speed of a road segment) to provide in the end something meaningful, while a model that consumes travel-time data will be more straight-forward}. 

Five main categories can be established to describe the spectrum of ITS data sources:

\begin{enumerate}
	\item Roadside sensing, which brings together tools and mechanisms that directly capture and convey data measurements from the road, obtaining valuable metrics such as speed, occupation, flow or even which vehicles are traversing a given road segment. These are the most commonly used sensors in ITS, most frequently based on computer vision and radar, as they directly provide traffic information close to the point where it originates. { This kind of sensed metrics are useful for traffic flow or speed modeling, allowing practitioners to identify mobility patterns and to model them, so future behavior in sensorized locations can be estimated. Counting vehicles or detecting their speed at a certain point of the road also allows to obtain network wide mobility patterns that can be compared to those provided by a simulation engine. This can help traffic managers and city planners take long-term decisions, such~as which road should be extended or how a road cut could affect other segments}. However, this~information is tethered to the exact points where the sensors are placed, thus~the actionability of a system built upon these data is subject to the geographical area where such sensing devices are deployed and their range.
	
	{
		\item In-vehicle sensing, which includes a broad range of transponder devices that are part of the on-board equipment of certain fleets. Commercial vehicles on land, air and water usually have location devices that record and emit the position and other metrics of the vehicles at all times. This opens up a wide range of ITS applications, such~as fleet management~\cite{said2016utilizing}, route optimization~\cite{urbahs2017remotely}, delay analysis and detection~\cite{khaksar2019airline}, airport/port management~\cite{mott2017estimating} or, when the vehicles are a part of traffic, a~detailed analysis of their behavior along a complete route (not only in certain sensorized locations)~\cite{herring2010estimating}. This technology is highly extended in commercial fleets and its multiple analytic applications are nowadays remarkably actionable, due to industry standards requirements. However, machine learning modeling based approaches are starting to emerge, and~should consider actionability as a core concern.
	}
	
	\item Cooperative sensing, which denotes the general family of data collection strategies that regards the information provided by different users of the ITS ecosystem as a whole, thus being grouped and jointly processed forward. This inner perspective of traffic and transportation can be obtained through many mechanisms, and, although it is more specific and scarce, it is also more complete than the one obtained from roadside sensing. These data open the door to mobility profiling and anomaly detection, enriching the outlook of a transportation model by means of the fusion of different data-based \emph{views} of an ITS scenario. This includes all forms of mobile sensing data, from~call detail record data that can be used to obtain users trajectories~\cite{kujala2016estimation}, to~GPS data~\cite{sun2015trajectory}. These sources are the foundation of abundant research~\cite{rodrigues2011mobile, lana2018road}, but in most cases the data fusion part is obviated. Crowdsourced and Social Media sensing can be analogously considered in this category. These data sources can also contribute to data-based ITS models by means of sentiment analysis and geolocation. The~use of crowd-sourced data is well established among technology-based companies (Google, Uber etc), yet not very often available to research community and private and public authorities in the transport operations management. The~limited information that becomes available is deprived from the necessary statistical representativeness and truthfulness in order to be easily integrated to legacy management systems.
	
	\item External data sources, which include all data that are not directly related to traffic of demand, but have an impact on it, such~as weather, calendar, or planned events, social and economic indicators, demographic characteristics etc. These data are usually easy to obtain, and~their incorporation to ITS models augments in general the quality of their produced insights and ultimately, the~actionability of the actions yielded therefrom. It is also true that this data source is typically unstructured, which can pose a challenge regarding its automatic integration.
	
	\item Structured/static data, which refers to data sources that provide information of elements that have a direct impact on transportation, such~as public transportation lines and timetables, or municipal bike rental services. Due to their inherently structured nature, data provided by these sources are often arranged in a fixed format, making it easier to incorporate to subsequent data-based modeling stages.{ Any of the previous data and applications can be enriched with these kind of data; a model that is able to represent the mobility of a city would probably enhance its capabilities if it considered these data. For~instance, a~bus timetable can help understand traffic in the street segments that are traversed by the bus service or where its stops are located.} These information sources must be considered for an intelligent transportation system to be actionable, being a particularly essential piece of urban and interurban mobility.
	
\end{enumerate}  

\subsection{Data Preprocessing}
The variety of the above mentioned sensing sources comes with promises and perils. These data is produced in various forms and formats, various time resolutions, synchronously or asynchronously and different rates of accumulation. To leverage the full spectrum of knowledge these data can bring to the sake of informed decision making, the~more the sensing opportunities the larger the needs for powerful preprocessing and skills are before reaching the stage of modeling.  	

A principled data-driven modeling workflow requires more than just applying off-the-shelf tools. In~this regard, preprocessing raw data is undoubtedly an elementary step of the modeling process~\cite{garcia2015data}, but still persists nowadays as a step frequently overlooked by researchers in the ITS field~\cite{lopes2010traffic}. 

To begin with, when a model is to be built on real ITS data, an important fact to be taken into account is the proneness of real environments to present missing or corrupted data due to many uncertain events that can affect the whole collection, transformation, transmission and storage process~\cite{vlahogianni2004short}. This issue needs to be assessed, controlled and suitably tackled before proceeding further with next stages of the processing pipeline. Otherwise, missing and/or corrupted instances within the captured data may distort severely the outcome of data-based models, hindering their practical utility~\cite{chen2001study}. A~wide extent of missing data imputation strategies can be found in literature~\cite{qu2009ppca,tan2013tensor}, as well as methods to identify, correct or discriminate abnormal data inputs~\cite{li2014missing}. However, they~are often loosely coupled to the rest of the modeling pipeline~\cite{ran2016tensor}. An actionable data preprocessing should focus not only on improving the quality of the captured data in terms of completeness and regularity, but also on providing valuable insights about the underlying phenomena yielding missing, corrupted and/or outlying data, along with their implications on modeling~\cite{lana2018imputation}.

Next, the~cleansed dataset can be engineered further to lie an enriched data substrate for the subsequent modeling~\cite{krempl2014open,etemad2018predicting}. A~number of operations can be applied to improve the way in which data are further processed along the chain. For~instance, data transformation methods can be applied for different purposes related to the representation and distribution of data (e.g.,~dimensionality reduction, standardization, normalization, discretization or binarization). Although these transformations are not mandatory in all cases, a~deep knowledge of what input data represent and how they contribute to modeling is a key aspect to be considered in this preprocessing stage. 

Furthermore, data enrichment can be held from two different perspectives that can be adopted depending on the characteristics of the dataset at this point. As such, feature selection/engineering refers to the implementation of methods to either discard irrelevant features for the modeling problem at hand, or to produce more valuable data descriptors by combining the original ones through different operations. Likewise, instance selection/generation implies a transformation of the original data in terms of the examples. Removing instances can be a straight solution for corrupted data and/or outliers, whereas the addition of synthetic instances can help train and validate models for which scarce real data instances are available. Besides, these approaches are among the most predominant techniques to cope with class imbalance~\cite{zheng2013using}, a~very frequent problem in predictive modeling with real data. Whether each of these operations is required or not depends entirely on the input data, their quality, abundance and the relations among them. This entails a deep understanding of both data and domain, which is not always a common ground among the ITS field practitioners~\cite{smith2004investigation}. 

Finally, data fusion embodies one of the most promising research fields for data-driven ITS~\cite{zhang2011data, el2011data}, yet remains marginally studied with respect to other modeling stages despite its potential to boost the actionability of the overall data-based model. Indeed, an ITS model can hardly be actionable if it does not exploit interactions among different data sources. Upon their availability, ITS models can be enriched by fusing diverse data sources. A~recent review on different operational aspects of data-driven ITS developments states that these models rarely count on more than one source of data~\cite{lana2018road}. This fact clearly unveils a niche of research when taking into account the increasing availability of data provided by the growing amount of sensors, devices and other data capturing mechanisms that are deployed in transportation networks, in~all sorts of vehicles, or even in personal devices held by the infrastructure users. Despite the relative scarcity of contributions dealing with this part of the data-based modeling workflow, the~combination of multiple sources of information has been proven to enrich the model output along different axis, from~accuracy to interpretability~\cite{choi2002data, chang2010intelligent,han2010radar, treiber2011reconstructing}. 

\subsection{Modeling}

Once data are obtained, fused, preprocessed and curated, the~modeling phase implies the extraction of knowledge by constructing a model to characterize the distribution of such data or their evolution in time. The~distillation of such knowledge can be performed for different purposes: to represent unsupervised data in a more valuable manner (as~in e.g.,~clustering or manifold learning), for instance, to insight patterns relating the input data to a set of supervised outputs (correspondingly, classification/regression) aiming to automatically label unseen data observations, to predict future values based on the previous values (time series forecasting), or to inspect the output produced by a model when processing input data (simulation). To do so, in~data-based modeling machine learning algorithms are often put to use, which allow automating the modeling process~itself. 

The above purposes can serve as a discrimination criterion for different algorithmic approaches for data-based modeling. However, when the goal is to model data interactions within complex systems such as transportation networks, it is often the case that the modeling choice resorts to ensembles of different learner types. For~instance, when applying regression models for road traffic forecasting, a~first clustering stage is often advisable to unveil typicalities in the historical traffic profiles and to feed them as priors for the subsequent predictive modeling~\cite{vlahogianni2009enhancing, lana2019, liu2019mining}. However, when it comes to model actionability, a~key feature of this stage is the \emph{generalization} of the developed model to unseen data. This~characteristic implies making a model useful beyond the data on which it is trained, which implies that the model design efforts should not only be put on making the model achieve a marginally superior performance, but also to be useful in other spatial or temporal circumstances. Achieving good generalization properties for the developed can be tackled by diverse means, which often depend on the modeling purpose at hand (e.g.,~cross-validation, regularization, or the use of ensembles in predictive modeling). Essentially, the~design goal is to find the trade-off between performance (through representing much of the intrinsic variance of data) and generalization (staying away from an overfitted model to a particular training set). This aspect becomes utterly relevant when data modeling is done on time-varying data produced by dynamic phenomena. ITS are, in~point of fact, complex scenarios subject to strong sources of non-stationarity, thereby calling for an utmost focus on this~aspect. 

The complexity met in traffic and transportation operations is usually treated with heterogeneous modeling approaches that aim to complement each other to improve accuracy~\cite{moretti2015urban,cong2016traffic,kim2015urban}. This can be done either by comparing different models and selecting the most appropriate one every time, or by combining different models to produce the final outcome. Additionally, in~some fields of ITS, such~as traffic modeling, physical (namely, theory- or simulation-based) models have been available for decades. Their integration into data-based modeling workflows, considering the knowledge they can provide, can~become crucial for a manifold of purposes, e.g.,~to enforce traffic theory awareness in models learned from ITS data. Indeed, the~hybridization of physical and data-based models has a yet to be developed potential that has only been timidly explored in some recent works~\cite{fusco2015short,montanino2015trajectory,chaulwar2016hybrid}.

Interestingly, complex data driven modeling solutions to transportation phenomena have been numerous and resourceful ranging from modular structures, to model combinations, surrogate modeling~\cite{vlahogianni2015optimization} and so on. Regardless of the approach, literature emphasizes on the critical issue of model hyperparameter optimization using for example nature inspired algorithms, namely Evolutionary Computation or Swarm Intelligence~\cite{cong2016traffic,teodorovic2008swarm}. Assuming that there is a feasible and acceptable solution to the problem of selecting the proposed parameters for a data drive model, when dealing with complex modeling structure this task should be conducted automatically by optimizing the hyperparameter space usually based on the models' predictive error. It is to note that, the~greater the number of models involved the more difficult the optimization task becomes. Moreover, relying on nature inspired stochastic approaches, full determinism in the solution and convergence stability can not be formally guaranteed~\cite{del2019bioinspired}. 

\subsection{Prescription}

Once the modeling phase itself has been completed, the~resulting model faces its application to a real ITS environment. It is at this stage when actions deriving from the data insights are defined/learned/decided, and~when the actionability of the model can be best assessed.{ Yet, this~stage is frequently overlooked in most ITS research, where most works conclude at presenting the good performance of a model; it is uncommon to find evaluations of a given model in terms of its final application in a certain environment. Are~the actions that can be taken as a result of the outcome of a data-based model aimed at a strategic, tactical or operational decision making? Is the output of the data-based model able to support decisions made by transportation networks managers? Can the output be consumed directly without any need for further modeling, or exploited by means of a secondary modeling process aimed at optimizing the decision making process?} This latter case can be exemplified, for instance, by the formulation of the decision making process as an optimization problem, in~which actions are represented by the variables compounding a solution to the problem, and~the output of the previous data-based modeling phase can be used to quantitatively estimate the quality or fitness of the solution. One of the most prominent examples of this prescription mode deals with routing problems, since they often use simulation tools or predictive models to assess the travel time, pollutant emissions or any other optimization objective characterizing the fitness of the tested routes~\cite{kumar2012survey,osaba2016improved}. Other examples of prescription based on data emerge in tactical and strategic planning, such~as the modification of public transportation lines~\cite{mendes2015validating}, the~establishment of special lanes (e.g.,~taxi, bike)~\cite{szeto2015sustainable}, the~improvement of road features~\cite{van2016automatic}, the~adaptive control of traffic signaling~\cite{mannion2016experimental}, the~identification of optimal delivery (or pickup) routes for different kinds of transportation services~\cite{osaba2017discrete}, the~incident detection and management~\cite{imprialou2013methods}, learning~for automated driving~\cite{yu2019distributed}, or the design of sustainable urban mobility plans based on the current and future demand or the drivers' behavior~\cite{lecue2014star,gindele2015learning}. 

In any of the above presented ITS cases, a~data-based model should be equipped with a certain set of features that guarantee its actionability. For~instance, if a traffic manager is not able to interpret a model or understand its outcome in terms of confidence, it~can be hardly applied for practical decision making. When the model is used for adaptive control purposes (as~in automated traffic light scheduling), the~adaptability of the model to contextual changes is a key requirement for prescribed actions to be matched to the current traffic status~\cite{kammoun2014adapt}. Interestingly, some control techniques with a long history in the field (e.g.,~Stochastic Model Predictive Control, SMPC,~\cite{mesbah2016stochastic}) serve as a good example of the triple-play between application requirements, decision making and data-based models. When dealing with the design of control methods in ITS, SMPC has been proven to perform efficiently in highly-complex systems subject to the probabilistic occurrence of uncertainties~\cite{hrovat2012development}. Specifically, SMPC leverages at its core data-based prediction modeling and low-complexity chance-constrained optimization to deal with control problems that impose that the method to be used must operate in real time. In~this case, and~in most actionable data-based workflows where decision making is formulated as an optimization problem, we~note a clear entanglement between application requirements (e.g.,~real-time processing), decision making (low-complexity, dynamic optimization techniques) and data-based models (predictive modeling for system dynamics forecasting). 
\subsection{Adaptation} \label{sec:adapt}

Finally, the~proposed actionable data processing workflow considers model adaptation as a processing layer that can be applied over different modeling stages along the pipeline. When models are based on data, they are subject to many kinds of uncertainties and non-stationarities that can affect all stages of the process. Streaming data initially used to build the model can experience long-term drifts (for instance, an increase of the average number of vehicles), sudden changes (a newly available road), or unexpected events (for example, a~public transportation strike)\cite{buchanan2015traffic, pan2013crowd,davison2006bus}. A~closed lane, a~new tram line, the~opening of a tunnel or simply the opening of a new commercial center, may change completely the way in which network users behave, and~thus, affect the data-based models that are intended to reflect such a mobility. Therefore, data-based modeling cannot be conceived as a static design process. This critical adaptation should be considered in all parts of the workflow, and~constantly updated with new data:

\begin{enumerate}
	\item In the preprocessing stage, adaptation could be understood from many perspectives: the incorporation of new sources of data, the~partial or total failure of data capturing sensors, which lead to an increased need for data fusion, imputation, engineering or~augmentation. 
	\item In the modeling stage, adaptations could range from model retraining, adaptation to new data or alternative model switching, to the change of the learning algorithm due to a change in the requested system requirements (for instance, in~terms of processing latency any other performance~indicator). 
	\item In the prescription stage, adaptation is intended to dynamically support decisions accounting for changes in data that propagate to the output of preceding modeling stages. Data-based modeling can deal with such changes and adapt their output accordingly, yet they are effective to a point. For~instance, online learning strategies devised to overcome from concept drift in data streams can speed up the learning process after the drift occurs (by e.g.,~diversity induction in ensembles or active learning mechanisms). Unfortunately, even when model adaptation is considered the performance of the adapted model degrades at different levels after the drift. Extending adaptation to the prescription stage provides an additional capacity of the overall workflow to adapt to changes, leveraging techniques from prescriptive analysis such as dynamic or stochastic optimization. 
\end{enumerate}

Adaptations within the above stages can be observed from two perspectives: automatic adaptations that the system is prepared to do when certain circumstances occur, or~adaptations that are derived from changes that are introduced by the user.~Thus, the~adaptation layer is strongly linked to actionability: an ITS model will be more actionable if adaptations, either needed or imposed, are accessible to its final users. For~instance, a~system could be required to introduce a new set of data, and~its impact on all the stages should be controlled by the transportation network manager, or if a drift is detected, the~system should consider if it is relevant to inform the user. 

\section{Functional Requirements for Model Actionability}\label{requirements}
\begin{figure}[H]
	\centering
	\includegraphics[width=0.80\columnwidth]{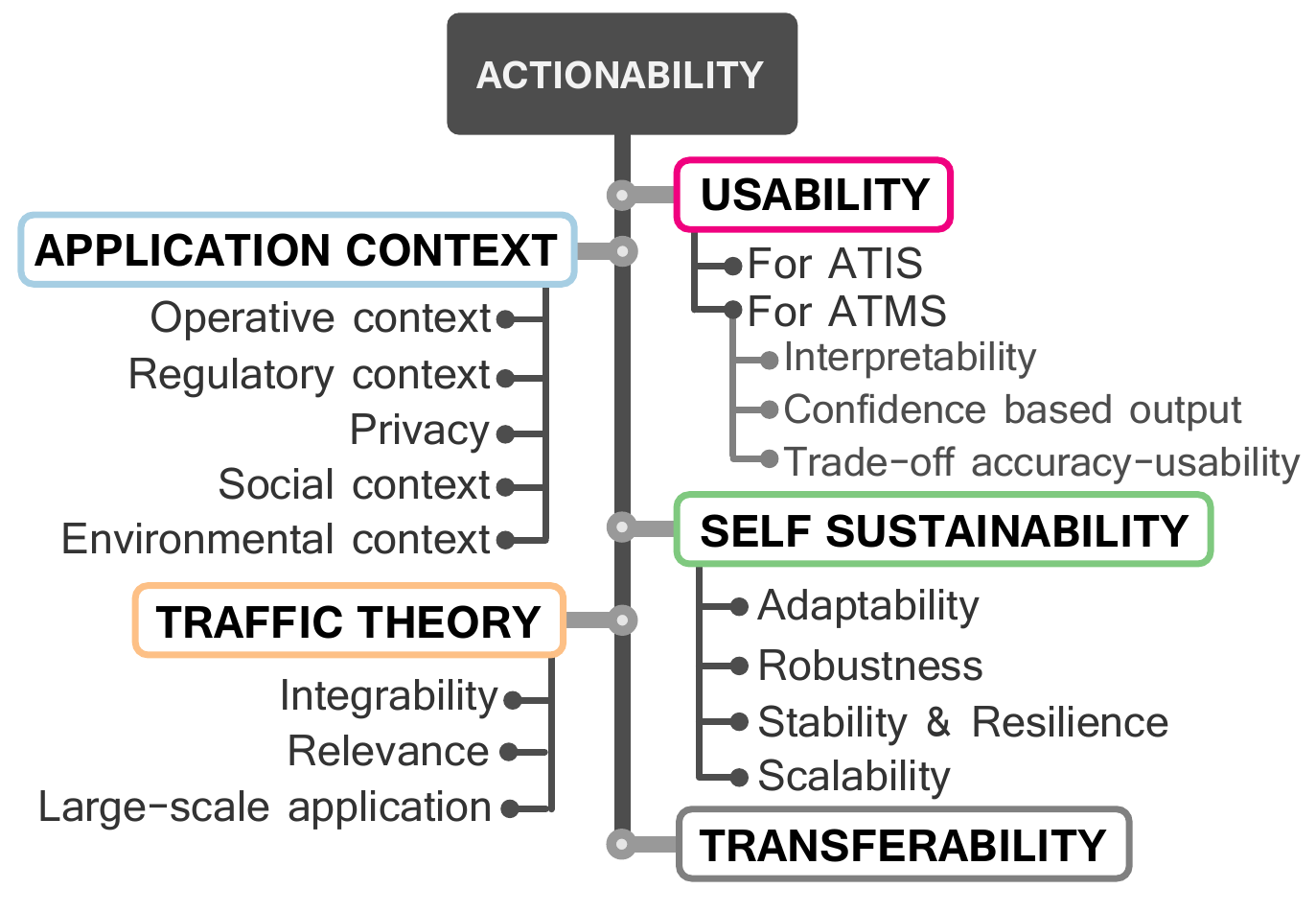}
	\caption{\vspace{0mm}Functional requirements for actionable data-based models in Intelligent Transportation Systems (ITS). ATIS: Advanced Traveler Information Systems; ATMS: Advanced Transportation Management Systems.} 
	\label{fig:2}
\end{figure}

Any data-based modeling process should embrace actionability as its most desirable feature for the engineered model to yield insights of practical value, so that field stakeholders can harness them in their decision making processes. This is certainly the case of ITS, in~which managers, transportation users and policy makers rely on models and research results to make better and more informed decisions. Thus, once the main stages of data-driven modeling have been outlined, this~section places the spotlight on the main functional features that should be mandatory to produce fully-actionable ITS data-based models. These functional requirements, which are shown in Figure \ref{fig:2}, should not be understood as a compulsory list of features, but rather as an enumeration of possibilities to make a model actionable. Not all ITS scenarios requiring actionable data-based models should impose all these requirements, nor can actionability be thought to be a Boolean property. Different loosely defined degrees of actionability may hold depending on the practicality of decisions stemming from the model.

\subsection{Usability}

The way in which humans interact with information systems has been thoroughly studied in last decades and formalized under the general \emph{usability} term~\cite{nielsen1994usability}. Although usability is a feature that can be associated to any system in which there is some kind of interaction with the user, most of its definitions to date gravitate around the design of software systems~\cite{nielsen1994usability2,brooke1996sus, nielsen199510}, which is not necessarily the case of ITS research. Usable designs imply defining a clear purpose for a system, and~helping users making use of it to reach their objectives~\cite{nielsen2003usability}. Within ITS, there are domains where this definitions apply directly~\cite{noy1997human}, such~as vehicle user interfaces~\cite{green1999estimating,green1999navigation,burns2010importance}, the~development of navigation systems~\cite{burnett2000turn}, road~signalization~\cite{dos2017proposal}, or even the way in which public transportation systems information is shown to users~\cite{avelar2006design,roberts2016radi}. 

The aforementioned domains of application, and~mostly any system lying at the core of Advanced Traveler Information Systems (ATIS), have an explicit interaction component. On the other hand, models developed for Advanced Transportation Management Systems (ATMS) are less related to user interaction (beyond the interface design of decision making tools), hence this canonical definition of usability seems to be less applicable. However, the~general concept of usability can also accommodate the notion of \emph{utility} as the quality of a system of being useful for its purpose, or the concept of \textit{effectiveness}, in~regards to how effective is the information provided by them~\cite{lyons2001advanced}. Since ITS are systems developed as tools designed to help the different stakeholders that take part in transportation activities, the~actionability of data-based models used for this regard depends stringently in this general idea of usability~\cite{barfield2014human}. 
Models' usability is a feature largely disregarded in literature. A~clear example of this situation is traffic forecasting, a~preeminent subfield of ATMS, in~which the link between the high end deep learning models with the requirements by the road operators in forecasts to support the decision making is very weak~\cite{Vlahogianni2014}. 

Usability may relate to the person that is going to operate the model, and~to the type and complexity of the model, which relate to specific skills. Achieving usable ITS models does not entail the same efforts for all ITS subdomains. Thus, while for research contributions related to ATIS there is a clear interest in this matter~\cite{horan2006assessing}, for ATMS developments some extra considerations need to be made. Usability in ITS has, therefore, a~facet oriented towards user interface, where interfaces reflect at least one of the outputs of an ITS data-based model, and~another facet towards creating models that are more aware of the way their outputs are going to be consumed afterwards by the decision maker. 

\subsubsection{User Interface}

For the first of these facets, Spyridakis~et~al.~\cite{barfield2014human} propose general software usability measuring tools and scales such as System Usability Scale (SUS)~\cite{brooke1996sus}, ethnographic field studies, or even questionnaires. These basic techniques are also proposed in~\cite{ross2001evaluating} in order to evaluate navigation systems interfaces. There are also many other evaluation measures that are more specific to the field, such~as~\cite{fischer2002human}, or those defined by public authorities~\cite{dingus1996development}. Some of the main techniques to appraise ITS interface usability are:
\begin{itemize}[leftmargin=*]
	\item Usability techniques: if the output of the developed model is consumed through the use of an interface, common techniques like asking directly the users about their experience can be adopted~\cite{ross2001evaluating}. Among them SUS surveys are the standard to provide interpretable metrics that can be used for the evaluation of passenger information systems~\cite{beul2014usability} or any other kind of automated traveler information system~\cite{horan2006assessing}.
	
	\item Quality of the provided information: in~\cite{lyons2001advanced}, another perspective is proposed, based on estimating the quality of the information provided by the model. Characteristics such as the means to access the information, the~reliability of the information provider, or the awareness of the information availability can be measured for assessing the model's usability.
	
	\item Transportation-aware strategies: an alternative way to measure usability is to take into account the transportation context and how the use of the model impacts the system. As many of these systems are used during the course of transportation, the~environment must be considered in order to provide an adequate and pertinent output~\cite{dingus1996development}. This particular aspect is regarded below in section \ref{sec:appcon}.
	
	\item Public transportation guidelines: when ITS developments are intended for the public domain, inclusion of disadvantaged collectives in the usability evaluation is a must~\cite{fischer2002human}. The~extent in which these concerns are addressed by the ITS solution should not be~disregarded.
\end{itemize} 

\subsubsection{Consumption of the Model's Output}

For this second usability facet, there are no scales or measurements in literature that provide an objective (or even subjective) usability assessments, but we propose some angles that should be considered when designing this kind of models: 

\begin{itemize}[leftmargin=*]
	\item\textit{Confidence-based outputs:}	data-driven models are often subject to stochasticity as a result of their learning procedure or the uncertainty/randomness of their input data (as~specially occurs in crowdsourced and Social Media data). This randomness imprints a certain degree of uncertainty in their outputs, which can be estimated values, predicted categories, solutions to an optimization problem or any other alike. Such~outcomes are often assessed in terms of their similarity to a ground truth in order to quantitatively assess the performance of the data-based model. Thus, a~practitioner aiming to make decisions based on the model's output is informed with a nominal performance score (which has been computed over test data), and~the predicted output for a given input. However, when one of such data-based models is intended to work in a real environment, there is no ground truth to evaluate the quality of the result they are providing towards making a decision.For instance, a~predictive model could score high on average as per the errors made during the testing phase. However, predictions produced by the model could be less reliable during peak hours than during the night, being less trustworthy in the first case as per the variability of the data from which it was learned, and/or the model's learning algorithm itself. For~this reason, the~estimation of the confidence of outputs from a data-based model must be analyzed for the sake of its usability. For~example, a~public transportation model that provides outlooks of future demand could be more usable if, besides the estimation itself, some kind of confidence metric was provided. Elaborating on this aspect is not very frequent in academic research, mainly due to the fact that confidence is not always that easy to obtain and the estimation procedure is, in~most cases, model-specific,requiring a previous statistical analysis of input data to properly understand their variability and characteristics. Unfortunately, such a confidence analysis is usually left out of the scope of research contributions, which rather focus on finding the best scoring model for a particular problem. Exceptions to this scarcity of related works are~\cite{mazloumi2011prediction}, in~which the uncertainty inherent to artificial neural networks is analyzed in a real ITS context; ~\cite{van2009bayesian}, in~which a committee of different models provides intervals of confidence to predictions;or the more recent contribution in~\cite{liu2019dynamic}, which~departs from previous findings in~\cite{tsekeris2009short, khosravi2011prediction} to estimate the uncertainty of traffic demand. This uncertainty estimation is then used as an input to assess the confidence of traffic demand predictions. These few references exemplify good practices that should be universally considered in contributions to appear.
	
	\item\textit{Interpretability:} a stream of work has been lately concentrated around the noted need for \emph{explaining} how complex models process input data and produce decisions/actions therefrom. Under the so-called XAI (eXplainable Artificial Intelligence) term, a~torrent of techniques have been reported lately to explain the rationale behind traditional black-box models, mainly built for prediction purposes~\cite{gunning2017explainable,arrieta2020explainable}. Nowadays, Deep Learning is arguably the family of data-driven models mostly targeted by XAI-related studies~\cite{samek2017explainable,ras2018explanation}.
	
	The interest of transport researchers to interpretable data-driven models is not new; intuitively, any decision in transportation and traffic operations should be based on a solid understanding of the mechanism by which different factors interact and influence transportation phenomena~\cite{vlahogianni2012modeling}. In~the transportation context explainability is closely related to integrability, when it comes to traffic managers, as ensuring that data-based models can be understood by non-AI expert can make them appropriately trust and favor the inclusion of data-based models in their decisional processes. When~framed within ITS systems and processes, the~need for explainable data-based models can help decision makers understand how information is processed along the data modeling pipeline, including the quantification of insightful managerial aspects such as the relationship and sensibility of a predicted output with respect to their inputs. 
	
	\item\textit{Trade-off between accuracy and usability:} when ITS data-based models aim at superior performances, they often work in ideal scenarios where the real context of application is disregarded; should that context apply in practice, the~claimed suitability of the developed model for its particular purpose could be compromised. For~instance, the~goodness of an ITS model devised to detect users' typical trajectories can be measured with regard to the exactitude of the detected trajectories. If the pursuit of a superb performance relies on a constant stream of data (hence, eventually depleting the user's phone battery), it could be a pointless achievement when put to practice. This particular example has been already considered by plenty of researchers~\cite{thiagarajan2009vtrack,thiagarajan2011probabilistic}. However, there is a long way to go in this aspect, as most ITS research developments consider only ideal circumstances without regarding the implications that an accurate design could have on its final usability.
\end{itemize}

\subsection{Self-Sustainability} \label{sec:self-sust}

In general, self-sustainability of a model refers to its ability to survive---hence, to continue to be useful---in a dynamic environment. ITS models and developments are usually intended to operate during long periods of time. However, it is widely accepted that traffic and transportation phenomena are strongly dynamic in nature, meaning that these phenomena exhibit long term trends, evolve in space and time, but also, at the occurrence of an unexpected event, they are susceptible to abrupt changes and exhibit long term memory effects. For~instance, a~trip information system based on traffic forecasts on a certain part of the network trained with historical data coming from recurrent traffic conditions may not be easily transferable to other road networks or not efficient in case of a severe disruption in traffic operation (accident). What is more, if the specific system does not undergo constant training with new data over time, eventually it will fail to correctly operate even for the network location it was originally designed to operate due to contextually induced non-stationarities. Thus, an intelligent transportation system developed based on data-based approaches should at least follows a set of minimum self-sustainability requirements during the design workflow. 

To better understand the importance of self-sustainability as a significant aspect of model's actionability, one should bring to mind the case of cooperative ITS systems (e.g.,~advanced vehicle control systems) and the automated driving. To this end, a~self-sustainable data-based model should bridge the gap between the development of a model prototype and its deployment in a real, potentially non-stationary environment. 

When an ITS system or model is deployed to operate in changing conditions, self-sustainability involves dealing with the effects of such changes in the learned knowledge. To this end, different strategies and design approaches could be required depending on the nature of the change and its effects on the model. We~next delve into several attributes that can be desirable to deploy data-driven systems or models in changing environments, rendering them actionable:
\begin{enumerate}[leftmargin=*]
	\item \textit{Adaptable}: Data-driven models for ITS applications created in controlled conditions, with static, self-contained datasets, can provide great performance metrics, but could also fail if data evolve along time~\cite{geisler2012evaluation}. Adaptation is the reaction of a system, model~or process to new circumstances intended to reduce its performance deterioration in comparison to the one expected before the change in the environment happened. If~data change over time, their evolution is not detected by the model and it does not adapt to it whatsoever, then the developed model will eventually provide an obsolete output. When these contextual variations occur over data streams and models are learned on-line therefrom (for e.g.,~on-line clustering or classification), such variations can imprint changes in the statistical distribution of input and/or output data, making it necessary to update such models to reflect this change in their learned knowledge. This phenomenon is known as \emph{concept drift}~\cite{gama2014survey}, and~has been identified as an active research challenge for most of fields connected to machine learning in non-stationary environment~\cite{vzliobaite2016overview}. Many of those fields are already studying this topic, from~spam detection~\cite{delany2005case,mendez2006tracking} to medicine~\cite{stiglic2011interpretability}. 
	
	There are two main lines related to concept drift: how to detect drift, and~how to adapt to it. Both lines should be scheduled in the research agenda of data-driven ITS, as they have obvious implications when analyzing traffic~\cite{moreira2014improving}. Situations like road works can modify completely traffic profiles over a certain area during a period of time, after which the situation goes back to normal. A~similar casuistry happens with road design changes (i.e.,~new lanes, transformation of types of lanes, new accesses, roundabouts, etc), although in those cases there is a new stable traffic profile largely after the change. Even without man-crafted changes, traffic profiles may change for social-economical reasons~\cite{lana2016role}. Besides, analysis of drift can be used to detect anomalies in the normal operation of roads~\cite{moreira2015drift3flow}, or to analyze patterns in maritime traffic flow data~\cite{osekowska2017maritime}. However, the~adaptability of ITS models to evolving data is scarcely found in literature, and~certainly, in~many cases concept drift management is the scope of the work, and~not a circumstance that is considered to achieve a greater goal~\cite{moreira2015drift3flow, wibisono2016traffic}. There are though some online approaches to typical ITS problems that consider the effects of drift in data~\cite{lana2019, wu2012online, procopio2009learning}, and~we consider this kind of initiatives should lead the way for an actionable ITS research. 
	
	\item \textit{Robust}: When an ITS system is deployed in a real-life environment, diverse kinds of setbacks can affect its normal operation, from~power failures that preclude its functioning to the interruption of the input data flow. Robustness is a self-sustainability trait that prevents a system to stop working when external disruptions occur. Although in most research-level designs this is not a relevant feature, it is essential for actionable, self-sustainable designs. Robustness, defined as the ability to recover from failures, would have, however, different requirements depending on the criticality of the ITS system. Thus, in~a traffic flow forecasting system robustness could only imply that the system does not crack when input data fail~\cite{zhang2008short}, and~it continues to operate; on~the other hand, for critical systems such as air traffic management, robustness would require additional measurements to contain damage~\cite{isaacson2010concept,chen2017air}. All in all, robust data-based workflows should be able to accommodate unseen operational circumstances, such~as data distribution shifts or unprecedented levels of information uncertainty, which particularly prevail in crowdsourced and Social Media data~\cite{wechsler2019pervasive, adar2007managing}.
	
	\item \textit{Stable and resilient}: Actionable systems require a certain output stability in order to be understandable by their users. This notion is apparently opposed to adaptability, but while the latter is the ability to adapt the output to environment or data changes, stability pursues maintaining the output statistically bounded even when contextual changes occur, through e.g.,~model adaptation techniques. When adaptation is not perfect and the model violates a given level of statistical stability, stability requires another kind of adaptation, namely \emph{resilience}, to make the model return to its normal operation and thus, minimize the impact of external changes on the quality of its output~\cite{de2017mathematical}. This entails, in~essence, going one step further in the knowledge of the environment and taking into account those circumstances that can affect the system, and~it could be linked to transferable models, which would be addressed below. For~instance, a~traffic volume characterization model would be adaptable if it considers the changes inherent to traffic volume (an increase over time due to economical factors), and~it would be stable if a change in the weather conditions does not deteriorate its performance, or in other words, it has considered this essential circumstance. These kind of considerations are almost nonexistent in literature~\cite{Vlahogianni2014}, but however crucial for a model to be self-sustainable. 
	
	\item \textit{Scalable}: In the research environment, tests are run in a delimited scale, constrained to the size of data, and~useful for the experiments, in~contrast with large, multi-variate real environments. Scaling up is not, of course, a~matter of ITS research, but~an engineering problem. However, models should be designed to be scalable since their~conception. 
	
	Leaving aside calibration and training phases, classic transportation theories tend in general to be computationally more affordable than data-driven models. However, the~unprecedented amount of computing power available nowadays discards any real pragmatic limitation due to the computational complexity of learning algorithms in data-based modeling. An exception occurs with models falling within the Deep Learning category which, depending on their architecture and size of training data, may require specialized computing hardware such as GPU or multi-core equipment. Nevertheless, the~rising trend in terms of scalability is to make data-based models incremental and adaptable~\cite{zhang2011data}, which finds its rationale not only in the environmental sustainability of data centers (lower energy consumption and thereby, carbon footprint), but also in the deployment of scalable model architectures on edge devices, usually with significantly less computing resources than data centers.
	
	Although some ITS problems are easier to scale and this feature would not be troublesome, there are some fields that can be very sensitive to scalability. For~instance, route~planners frequently consist of shortest-path problem and travel-salesman problem implementations that increase in complexity when the number of nodes grow~\cite{colpaert2016impact}. This is a good example where artificial intelligence and optimization tools provide solutions that are actionable in terms of scalability, and~where cases are found effortlessly~\cite{basu2016genetic,schmitt2018experimental}. Caring about aspects like the easiness to introduce new variables when needed, the~complexity of tuning if applies, or the execution time, would make a model more actionable, by increasing its self-sustainability. This need for scalability is not just a matter related to the computational complexity of modeling elements along the pipeline, but also links to the feasibility of migrating the designed models from a lab setup to a, e.g., Big Data computing architecture. Unfortunately, scarce~publications reflect nowadays on whether their proposed data-based workflows can be deployed and run on legacy ITS systems, thereby avoiding costly upgrade investments in computing equipment.
\end{enumerate}

\subsection{Traffic Theory Awareness}
Theoretical representations of traffic attempt to construct (mostly simple) models with causal aspects. These models are usually of a closed form and are frequently dictated by simplifying assumptions, which leads to limited performance when modeling complex spatio-temporal dynamics in the microscopic analysis context. In~these models, data are instrumental to estimate how well they fit real world conditions. On the other hand, and~since their upsurge in the 80s, data driven models rely exclusively on the data to extract the dynamics that govern the phenomena. This, at least theoretically, makes them more adaptable and more efficient in complex conditions when compared to theory based models. But, they can hardly claim applicability in large scale scenarios (city level traffic management) due to significant computational resources requirements. Such data-driven traffic models have been systematically implemented as proof of concepts and are now dominant in Traffic Engineering literature~\cite{lana2018road}, incorporating most well-known advanced techniques, and, in~many cases, ignoring the elementary knowledge of traffic and focusing blindly on performance.

Owing to the above, researchers in traffic modeling have diversified the way in which their models are developed and evaluated, fitting them to the technology that is introduced, as opposed to fitting the model to the well established knowledge described in well established theories of traffic flow. This results in models that are hardly actionable for traffic engineers, in~terms of integration to legacy traffic control and management systems and relevance to the decision making process of road network operators. Besides, there is a lack of standards in what regards to data and scenarios used to assess the performance, usually due to the availability of real data for each researcher. This was already identified in~\cite{Vlahogianni2014}, where test-beds were proposed, either generating them or using some of the existent as standards. This would help compare models, understand them better, as they can be evaluated in a known environment, and~obtain their insights concerning traffic theory. Besides, as we anticipated in Section~\ref{sensing} there is a industrial trend towards the consideration of different data sources when modeling traffic dynamics. In~many cases, these data sources do not have any straightforward relationship to traffic itself. The~integration of these sources of data, the~models learned from them and theoretical representations of transportation scenarios remains an open challenge that has started to be addressed in literature~\cite{zhang2012exploring, zhang2016exploratory, zhang2017understanding}. 

In this line of reasoning, linking data-driven to theory based models in transportation may resort to efficient and physically consistent representations of transportation phenomena. In~fields like traffic modeling and forecasting, this~hybrid approach permits to consider theoretical aspects of traffic, such~as the relationship among speed, flow and density, the~three phases of traffic~\cite{kerner2004three}, or the Breakdown Minimization Principle~\cite{kerner1999physics} when modeling bottlenecks. The~consideration of these theoretical concepts takes effect mainly in the preprocessing, modeling and prescription phases of the modeling workflow. In~preprocessing, domain knowledge can be crucial for feature engineering, by describing how available features are related to each other, estimating collinearities in advance, deleting irrelevant predictors, or obtaining feature combinations with improved modeling power ~\cite{Vlahogianni2014}. Applying traffic theories and principles can also be useful for data augmentation and missing data imputation, by simulating or generating data that are more akin to what the context can provide ~\cite{lana2019}. In~the modeling phase, previously defined mathematical frameworks can help define the constraints, operation ranges and correct the output of data-based models, which do not take into account the compliance of their output with respect to well-established theories. Lastly, in~the prescription phase, model outputs can be linked to traffic theory knowledge to improve the way in which they are applied: a~predicted flow value can be more useful if the travel time or the bottleneck probability can be computed afterwards. Furthermore, in~the case of predictive models, they can reach a point in which the provided predictions ultimately affect the future behavior of the models themselves, if they are trained only with observed past data. For~instance, a~model that assists traffic management decisions, like closing a lane, might lead to a situation that has not been observed by the model before, thus making the knowledge captured by the traffic model obsolete and useless until the data captured from the environment is exploited for retraining. Physical models can be highly useful to anticipate scenarios and complement data-based models, providing additional information of what theories or simulations determine that the behavior of the scenario should be.

This emergent modeling paradigm is known as Theory Guided Data Science, and~aims to enhance data driven models by integrating scientific knowledge~\cite{karpatne2017theory}. The~main objective of this approach is to enable an insightful learning using theoretic foundations of a specific discipline to tackle the problem of data representativeness, spurious patterns found in datasets, as well as providing physically inconsistent solutions. From the algorithmic point of view, this~induction of domain knowledge can be done in assorted means, such~as the use of specially devised regularization terms in predictive models (e.g.,~in the loss function of Deep Learning models), data cleansing strategies that account for known data correlations, or memetic solvers that incorporate local search methods embedding problem-specific heuristics. In~transportation, there has been several example of theory enhanced models departing from traffic conditions identification and characterization~\cite{vlahogianni2007spatio,ramezani2012estimation}, to~data driven and agent based traffic simulation models for control and management~\mbox{\cite{zhang2011data,chen2010review,montanino2015trajectory,shahrbabaki2018data},} or cooperative intelligent driving services~\cite{mintsis2017evaluation}.

Awareness with domain-specific knowledge can be also enforced at the end of the workflow. When decision making is formulated as an optimization problem, the~family of optimization strategies known as Memetic Computing~\cite{gupta2018memetic,neri2012memetic} has been used for years to incorporate local search strategies compounded by global search techniques and low-level local search heuristics. These heuristics can be driven by intuition when tackling the optimization problem at hand or, more suitably for actionability purposes, by a priori knowledge about the decision making process gained as a result of human experience or prevailing theories. For~instance, traffic management under incidences in the road network can largely benefit from the human knowledge acquired for years by the manager in charge, since this knowledge may embed features of the traffic dynamics that are not easily observable from historical data. This knowledge can be inserted in an optimization algorithm devised to decide e.g.,~which lanes should be rerouted in an accident.

\subsection{Application Context Awareness} \label{sec:appcon}

Transportation is exceptionally diverse around the world, with notable differences in modes, preferences and availability due to social, economic and cultural disparity. Moreover, Intelligent Transportation Systems with different purposes have also characteristic requirements that can also be very divergent with respect to space and time. To address this landscape of complex and some times conflicting goals, policies and decision making should span from few seconds (traffic management and control) to years (planning and designing of new systems). It is strongly argued that data driven framework are able to cope with context aware datasets, due to their inherent capabilities of learning patterns hindering in resourceful data and reconstruct - in a sense - the context of the application. Typical examples of such context aware systems are the extraction of Origin-Destination matrices from cellphone based data~\cite{gonzalez2008understanding}, the~mobility applications that aim to improve the the mobility footprint of users~\cite{chatterjee2018type2motion}, as well as the smartphone based driving insurance systems~\cite{tselentis2017innovative}. Although these approaches seem to be appropriate to complement the user or system's experience on a problem, significant uncertainty lies in their transferability and accuracy, owing to the lack of context-aware knowledge.  

A certain degree of awareness of the context should be a matter of concern when developing ITS models that intend to be actionable. Context aware information is usually introduced in the modeling, for example accounting fro the demographic characteristics of the application area, the~type of the road or network, the~mode, the~travel purpose etc. However, what is usually disregarded is a much broader consideration of the operational and system's characteristics, such~as how models can be introduced to the operations at hand, what the privacy concerns are with respect to data and information flows, what is the regulatory framework and policy level restrictions and goals to be reached. 

First, within the operation, the~deployment context where a developed model is intended to be implemented can enforce a series of operative constraints. Creating and proposing an ITS model without observing these requirements is an exercise of futility, for its lack of actionability. From this operation perspective, the~context covers from deployment and operation costs---is the system cost-efficient considering its potential service?---to functioning modes---has the model the expected response times? can it operate in reduced computational power environments? As an illustration, a~system designed to detect and identify pedestrians can be very effective in terms of performance, but if it does not operate at an appropriate speed, or it needs more demanding computations that cannot be embarked in a vehicle, it is useless for an autonomous driving context~\cite{andreasson2015autonomous}. A~similar reasoning holds if by \emph{operation cost} one thinks about the energy consumption of the model at hand. Questions such as whether the energy consumed by the model compliant with the system should be kept in mind at design time, but also from the academic perspective, where efforts should be directed to the development of models that are consequent with the actual operative circumscription.

Second, regulations constitute a hard and highly contextual constraint in the implementation of ITS. Besides the wide regulatory differences that can be found across regions, there are transport frameworks where regulations are specially rigid. A~typical example is the case of airports~\cite{kulik2016intelligent}, and~where there is a broad field for specialized ITS. Another example is the constantly rising use of drone systems to monitor traffic~\cite{barmpounakis2016unmanned}. Models that fail to relate to the application's regulatory environment are not actionable. 

Third, data privacy and sovereignty constitute a growing concern in a connected world where, after a decade of handing over data with complacency, an awareness about personal information sharing is springing. A~recent example is the introduction of the EU General Data Protection Regulation (GDPR) framework, that severely disputed the manner data were introduced to models, as well as data availability. ITS models that are based on personal data are common nowadays, for instance in floating car data based developments~\cite{lin2014mining}. However, there are fields where this aspect is becoming crucial (autonomous driving connectivity~\cite{khodaei2015key}, security in public transport environments~\cite{menouar2017uav}), and~research is steering to privacy-preserving approaches~\cite{sucasas2016autonomous}, spheres where technologies such as Blockchain can have a major dominance~\cite{yuan2016towards,lei2017blockchain}.

Fourth, social aspects of the application play a major role in modeling. Social transportation is the subfield in ITS where the ``social'' information coming from mobile devices, wearable devices and social media is used for a number of ITS management related applications~\cite{zheng2015big}. The~outcomes from social transportation may be, to name a few, traffic analysis and forecasting~\cite{he2013improving,ni2016forecasting}, transportation based social media~\cite{evans2012microparticipation}, transportation knowledge automation in the form of recommending systems and decision support systems~\cite{kuflik2017automating}, and~services for the collection of further signal to be used later for the already mentioned purposes or others. However, cultural differences can have a relevant impact in how these systems operate, as social data are most commonly strongly linked to geographical information. This is a key aspect for their actionability. 

Fifth, transportation is currently a large source of greenhouse-gas emissions~\cite{woodcock2009public}. These concerns are gaining momentum in a wide range of ITS applications, such~as the discovery of parking spots~\cite{chen2013development}, multimodality applications that grant travelers the chance of using collective transportation systems efficiently and conveniently~\cite{kramers2014designing}, the~improvement of logistics operations~\cite{zhang2015swarm}, shared mobility applications, which help reducing the number of one-passenger vehicles in the road network~\cite{feigon2016shared}, or driving analytics to improve safety and ecological footprint~\cite{vlahogianni2017driving, huang2018eco, adamidis2019impacts}.  

Of course, research goes beyond the application context and does not need to be always connected to a certain application scenario. A~prototype can be far from the practical requirements of its eventual deployment; still, knowing the essential application common grounds is key to converge to actionable models. Unfortunately, this~is a matter frequently disregarded in ITS research.

\subsection{Transferability}

Within the research context, it is common to employ test data to assess the models. Regardless if these data are obtained from real sources or synthetically generated, the~resulting models have been built around them, and~can be heavily linked to that experimentation context. Would these models work in other context or with other input data? Transferability could be defined as the quality of a data-driven model to be applied in other environment with other data, and~it is directly linked with actionability: the application of a model should be generalizable to different datasets and transportation settings. This definition stems from the more general concept of \emph{Transfer Learning}~\cite{pan2009survey}, which can entail that models trained in a certain domain are applied to other domains, so that the previous knowledge obtained from the first makes them perform better in the latter than models without it. 

Depending on the subcategory of ITS, this~requirement can be easily met or arduous to achieve, as some subcategories are more oriented towards the application and rely less on the environment than others; the key is defining what is \emph{environment}. For~example, a~travel time forecasting model developed with data of a certain location could be transferable to another location without great complications, if it is built considering this feature~\cite{bajwa2005performance}. In~fact, many ITS models that are spatial-sensitive are developed using real data, but within the experimentation context, they are evaluated only in certain locations. Transferability for these scenarios would imply that the obtained results are reproducible (with certain degree of tolerance) in other locations.This could entail from plainly extrapolating the model to other locations~\cite{getachew2007simplified}, to implementing of techniques such as \emph{soft-sensing}, aimed at modeling situations where no sensor is available~\cite{habibzadeh2018soft}, and~the environment information is enough to obtain these models. A~similar case in terms of spatial contexts, but with more parameter complexity, requires plenty of information about the environment. As an illustration, the~case of crash risk estimation implies a higher calibration and adjustment needs due to the higher number of parameters that take part in this type of estimations. In~these circumstances, works such as~\cite{shew2013transferability} or~\cite{xu2014using} work with posterior probability models and give more relevance to models that behave with a certain performance in many contexts than to models that perform better in a particular location. On the other extreme, for cases like autonomous driving, the~change of environment is connatural to the domain (a moving vehicle constantly changing its location), and~the parameters of these models are abundant and highly variable. Thus, these applications need transferable solutions, transferability that is specifically sought by researchers, for instance in LIDAR based localization~\cite{ibisch2013towards} or pedestrian motion estimation~\cite{shen2018transferable}. In~any case, and~regardless the domain, ITS research is in an incipient stage (probably with the exception of autonomous driving) of developing transferable models, and~evaluating this feature, and~some machine learning paradigms can help improve this characteristic.

\section{Emerging AI Areas towards Actionable ITS}\label{challenges}

We have hitherto elaborated on the requisites that a model should meet towards leading to actionable data-based insights in ITS applications and processes. Some of these requirements can be fulfilled by properly designing the data-based workflow (e.g.,~interpretability can be straightforward for certain prediction models, whereas adaptability can be enforced by periodically scheduling the learning algorithm under use and feeding it with new data). However, several research areas have stemmed in the last years from the wide fields of Data Science and Artificial Intelligence that may serve to catalyze the compliance of data-based ITS workflows with the prescribed requisites, and~thereby attain the sought actionability of their produced insights.

The main AI areas that have been identified as potentially appropriate for addressing the requirements can be summarized briefly as follows:
\begin{itemize}[leftmargin=*]
	\item Real-time data processing and online learning, which are not brand new research avenues in ITS, as we can find advanced developments in the literature. However, as we will later show, emerging fields with great potential such as dynamic data fusion and dynamic optimization can expedite and proliferate the adoption of incremental data-based models in more ITS-related applications.
	
	\item Transfer learning and domain adaptation, that could allow to develop models for certain contexts and export them to others, linking directly to the transferability requirement, but also to  the integration of transportation theories and physical models to data-based models.
	
	\item Gray-box modeling, a~paradigm halfway between white-box (physical) and black-box (data-based) models. Gray-box modeling represents a promising area to bring awareness to traffic theory and other physical modeling when developing data-based models, with the potential to increase the performance, usability and comprehensibility of the latter. 
	
	\item Green AI, a~trend in Artificial Intelligence research that connects directly with energy and cost efficiency. Developing efficient models has a relevant impact in their sustainability and context awareness.
	
	\item Fairness, Accountability, Transparency and Ethics: Data-based models---specially those learning from large amounts of diverse data from many sources---are fragile to biases, and~can compromise aspects such as the fairness of decisions or the differential privacy of data. In~this context of growing sources of data, including those gathered from people, and~increasingly opaque data-based models, it has become essential to understand what models have learned from data, and~to analyze them beyond their predictive performance to consider ethical, societal and legal aspects. These aspects have been scarcely considered in ITS research.
	
	\item Other Artificial Intelligence areas such as imbalanced learning, reinforcement learning, adversarial machine learning are later highlighted for their noted relevance in ITS.
\end{itemize}

\begin{figure}[H]
\includegraphics[width=.95\columnwidth]{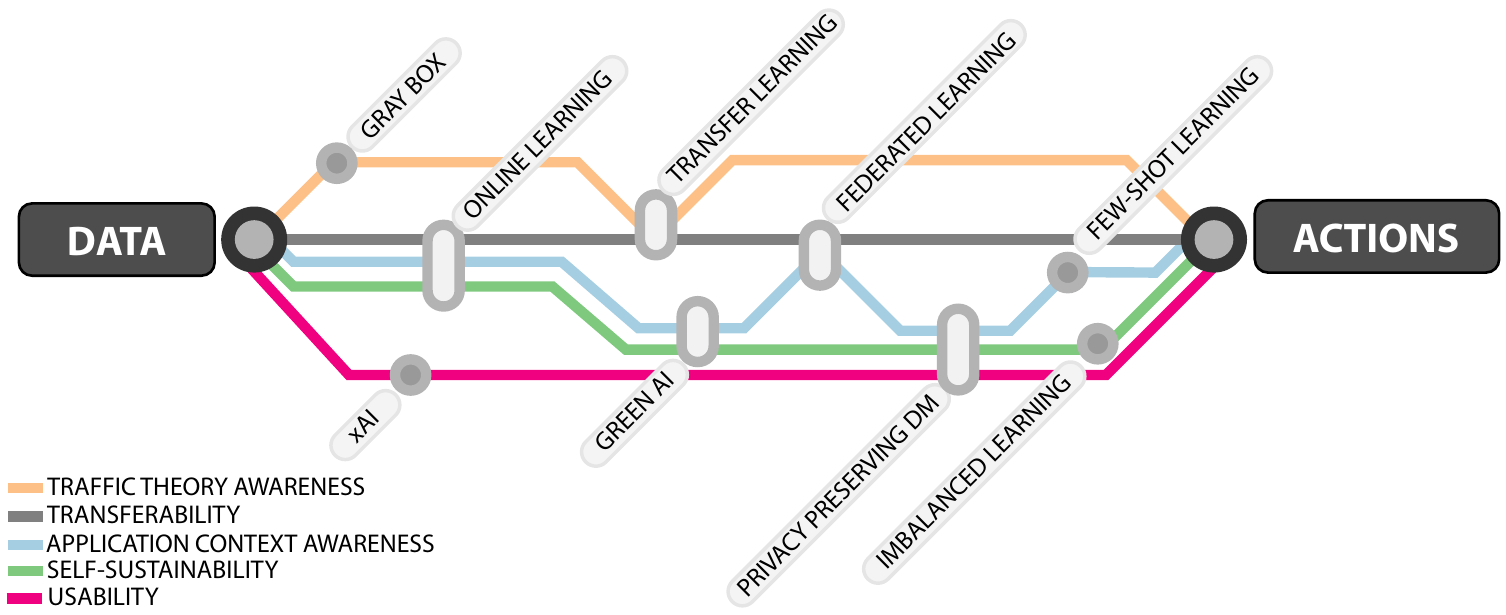}
\caption{Schematic diagram showing how avant-garde AI subareas can promote actionability in ITS data-based modeling workflows. Subareas contributing with particular emphasis to different functional requirements are connected together along the way from data to actions.}
\label{fig:3}
\end{figure}

We next discuss on the research opportunities spurred by the above research lines, their connections with the requirements presented in Section~\ref{requirements} (shown in Figure \ref{fig:3}), as well as the challenges that stem from the consideration of these AI areas in the context of ITS.

\subsection{Online Learning and Dynamic Data Fusion/Optimization}

Previously sketched in Section~\ref{sec:self-sust}, by online learning we refer to the capability of the learning model and in general, of the entire workflow, to learn from fastly arriving data possibly produced by non-stationary phenomena that enforces a need for adapting the knowledge captured by the model along time. Changes over data streams can make the data pipeline obsolete, thus demanding active or passive techniques to update it with the characteristics of the stream~\cite{ditzler2015learning,gama2014survey}. 

Although activity around online learning has mostly revolved on certain clustering and classification paradigms (the latter giving rise to the so-called concept drift term to refer to pattern changes), it is important to note that adaptation can be also needed in other stages of the actionable data-based workflow, from~data fusion to the prescription of actions. This~being said, research areas such as dynamic optimization and dynamic multi-sensor data fusion should be also investigated deeply in future studies related to actionable data-based models, specially when the scenario under analysis can produce information with non-stationary statistical characteristics. When merging different data sources, fusion strategies at different levels can be designed and implemented, from~traditional means (data-level fusion, knowledge-level fusion) to modern methods (corr. model-based fusion, federated learning or multiview learning)~\cite{smirnov2019knowledge,wang2019data}. Fusion of correlated data sources can compensate for missing entries or noisy instances in static environments. However, when data evolve over time as a result of their non-stationarity, new challenges may arise in regards to the inconsistency among multiple information sources, including measurement discrepancy, inconsistent spatial and temporal resolutions, or the timeliness/obsolescence of the data flows to be merged, among other issues. For~this reason, close attention should be paid to advances reported around adaptive fusion methods capable of detecting, counteracting and correcting misalignments between data flows that occur and evolve over time. This~branch of dynamic data fusion schemes aims at combining together information flows produced by non-stationary sources, synthesizing a representation of the recent history of each of the flows to be merged into a set of more coherent, useful data inputs to the rest of the data-based pipeline~\cite{khaleghi2013multisensor,ramachandran2006dynamic}. On the other hand, dynamic optimization techniques can efficiently deliver optimized actionable policies when the objectives and/or constraints of the underlying optimization problem varies~\cite{nguyen2012evolutionary,mavrovouniotis2017survey}. We~energetically advocate for a widespread embrace of advances in these fields by the ITS community, emphasizing on those scenarios whose dynamic nature can make the obtained actionable insights eventually obsolete. This is the case, for instance, of traffic related modeling problems (e.g.,~traffic flow forecasting and optimal routing) or driver characterization for consumption minimization, among many others.

Other requirements for actionability can also benefit from the adoption of the above models in dynamic ITS contexts. For~instance, cost efficiency in terms of energy consumption can largely harness the incrementality that often features an online learning model. The~use of dynamic data fusion can also yield a drastically less usage of communication resources in wireless V2V links, such~as those established in cooperative driving scenarios. All in all, the~recent literature poses no question around the relevance of adaptation in data-based modeling exercises noted in this work, with an increasing volume of contributions dealing with the extrapolation of adaptation mechanisms to ITS problems~\cite{chang2009online,saadallah2018bright,moreira2013predicting}.

\subsection{Transfer Learning and Domain Adaptation} \label{sec:tfada}

In close semantics to its related actionability requirement (\emph{transferability}), transfer learning aims at deriving novel means to export the knowledge captured by a data-based model for a given task to another task with different inputs and/or outputs~\cite{pan2009survey}. Depending on the amount of alikeness between the origin task and the destination task, we~may be also referring to \emph{domain adaptation}, by which we adapt the model built to perform a certain task to make it generalize better when processing new unseen inputs that do not follow the same distribution as their original counterparts (only the distribution changes~\cite{sun2015survey}). Techniques such as subspace mapping, representation learning, of feature weighting arise as those methods most used to allow knowledge to be transferred between data-based models used for prediction. 

In essence, transfer learning can provide higher prediction accuracy for models whose number of parameters to be learned (e.g.,~weights in a Neural Network) demands higher amounts of labeled data than those available in practice. However, data augmentation is not the only goal targeted by transfer learning. Domain adaptation may yield a better performance when used between ITS models that can become severely affected by a lack of calibration, different configurations or diverging specifications. An immediate example illustrating this hypothesis is the use of camera sensors for vehicular perception. Models trained to detect and identify objects in the surroundings of the vehicle can fail if the images provided as their inputs are produced by image sensors with new specs. The~same holds for car engine prognosis: replaced components can make a data-based characterization of the normal operation of the engine be of no practical use unless a domain adaptation mechanism is applied. Personalization of ITS services can be another problem where domain adaptation can help refine a model trained with data from many sources: a clear example springs from naturalistic driving, where a behavioral characterization model built at first instance from driving data produced by many individuals (source domain) can be progressively specialized to the particular driver of the car where it is deployed~\cite{ou2018transfer,xing2019driver,xing2018end}.

In regards to actionability, several functional requisites can be approached by using elements from Transfer Learning over the data-based pipeline. To begin with, it should be clear that the transferability of learned models for their deployment in different locations and contexts could be vastly improved by Transfer Learning, as the purpose of this AI branch is indeed to meet this requirement in data-based learning models. In~fact, this~approach is currently under study and wide adoption within the ITS community working on vehicular perception: when the capability of the vehicle to sense and identify its surrounding hinges on learning models (e.g.,~Deep Learning for image segmentation with cameras), a~plethora of contributions depart from pretrained models, which are later particularized for the problem/scenario at hand~\cite{ye2018machine}. This exemplified use case supports our advocacy for further efforts to incorporate transfer learning methods in other ITS applications, specially those where data collection and supervision are not straightforward to achieve in practice. Another functional requirement where Transfer Learning can pose a difference in ITS developments to come is cost efficiency. The~knowledge transferred between models learned from different contexts can improve their performance, thereby reducing the need for supervising data instances and ultimately, the~time, costs and resources required to perform the data annotation. 

Finally, the~more recent paradigm coined as Federated Learning refers to the privacy-preserving exchange of captured knowledge among models deployed in different contexts~\cite{konevcny2016federated,mcmahan2017communication}. Although the main motivation for the initial inception of Federated Learning targeted the mobile sector, techniques supporting the federation of distributed data-based models can be of utmost importance in the future of ITS, specially for V2V communications among autonomous vehicles and in-vehicle ATIS systems. Definitely the enrichment of models with global knowledge about the data-based task(s) at hand will pose a differential breakthrough in vehicular safety and driving experience. For~instance, federated models can collectively identify, assess and countermeasure the risk of more complex vehicular scenarios than each of them in isolation~\cite{ferdowsi2019deep}. Likewise, ATIS systems can learn from the preferences and habits of other users to better anticipate the preferences of the driver and act accordingly~\cite{vogel2018emotion}. In~a few words: an enhanced and more effective actionability of the data-based workflows built to undertake such tasks.

\subsection{Gray-Box Modeling}

Gray-box modelling refers to the design of models that combine theoretical developments and structures related to the problem, with data that serve as a complement for such theories to make the overall model match better the scenario under analysis~\cite{kroll2000grey,oussar2001gray}. Gray-box models lie in between white-box models, for which the learned structure is deterministic and grounded in theoretical concepts; and black-box models, 
whose~internal structure lacks physical significance and is learned from data. An example of white-box model in ITS systems is the use of computational fluid dynamics for macroscopic traffic flow modeling, whereas Deep Learning models for traffic forecasting can exemplify black-box modeling in this domain. Gray-box models have been lately embraced by the ITS community in a number of modeling scenarios, such~as those combining biological concepts and data-based models for driver characterization~\cite{inga2015gray,flad2017cooperative}.

Gray-box modeling can contribute to the actionability of data-based workflows for ITS applications in two different albeit interconnected directions. To begin with, the~incorporation of theoretical models to data-based pipelines can narrow the gap between engineers and practitioners more acquainted with traditional tools to analyze ITS systems and processes. Indeed, hybrid modeling can tie both worlds together not only without questioning the validity of prevalent theoretical developments, but also evincing the complementarity and synergy of both approaches. On the other hand, using validated theoretical models can help data-based modeling overcome difficult learning contexts such as class imbalance, outlier characterization or the partial interpretability of data clusters, among others. 

\subsection{Green Artificial Intelligence}

A profitable strand of literature has recently stressed on the energy efficiency of data-based models, highlighting the need for redesigning their learning algorithms to minimize their energy consumption and thereby, make them implementable and usable in practice~\cite{mittal2016survey,alwadi2015energy,han2013approximate}. While this issue is particularly relevant for resource-constrained devices (e.g.,~mobile hand-helds), the~concern with energy efficiency goes beyond usability towards environmental friendliness. For~this reason many recent contributions are striving for computationally lightweight variants of machine learning models that sacrifice performance for a notable reduction of their energy demand. This is not only the case of predictive models capable of incrementally learning from data, but also of specific Deep Learning architectures tailored for their deployment on embedded devices~\cite{lane2015can}.

Based on the above rationale, cost efficiency is arguably the most evident functional requirement around which energy-aware model designs can pose a breakthrough towards improving the actionability of the overall data-based workflow. In~addition, other aspects can be made more actionable by using energy-aware model designs, such~as usability~\cite{faisal2015towards}. Despite achieving unprecedented levels of predictive accuracy, a~data-based workflow may become useless should it deplete the battery of the system on which it is deployed for operation. Therefore, energy efficiency should be under the target of future research efforts, specially when dealing with ITS applications running on battery-powered devices, inspecting interesting paths rooted thereon such as the trade-off between performance and energy consumption, or the adaptation of the model's operation regime depending on the remaining battery life, among others~\cite{zliobaite2015towards}.

\subsection{Fairness, Accountability, Transparency and Ethics}

To end with, the~prescription of actions based on the insights provided by a data-based pipeline must be buttressed by a thorough understanding of the mechanisms behind its provided decisions~\cite{1910.10045}. Extended information about the model must be presented to the end user for several reasons: 
\begin{itemize}[leftmargin=*]
	\item To gauge as many consequences of the actions as possible, identifying situations where decision making based on the outputs of the data-based workflow gives rise to socially unfair scenarios due to the propagation of inadvertently encoding bias to the automated decisions of the model.
	\item To ensure him/her that the output of the model is reliable and invariant under the same data stimuli, maintaining a record of the intermediate decisions made along the pipeline, allowing for the post-mortem, potentially correcting analysis of bad decision paths, and~thereby maximizing the trust and certainty of the user when embracing its~output.
	\item To make the user understand why the developed model produces its prescriptive output when fed with a set of data inputs, shedding light on which inputs correlate more significantly with the prescribed actions, tracing back causal relations between intermediate data inputs, and~discriminating extreme cases where decisions can change radically under slight modifications of the model inputs. 
	\item To supervise the ethics of data-based workflows, identifying potentially illegal uses of unlawful data given the prevailing legislation, guaranteeing the privacy and governance of personal data by third-party data-based ITS applications and processes, and~certifying that the output of the model's output does not favor inequalities in terms of gender, religion, race or any other aspect alike.
\end{itemize}

The above requirements have been lately collectively compiled under the FATE (Fairness, Accountability, Transparency and Ethics) concept, which refers to the design of actionable data-based pipelines whose internal operations can be explained, accounted and critically examined in regards to the consequences of their eventual bias in privacy, fairness and ethical issues~\cite{martin2018ethical,veale2017fairer,stoyanovich2017fides}. This recent concern with the operation of machine learning models spawns from the proliferation of real cases where practical model installments have unveiled deficiencies of different kind, from~differential privacy breaks (data revealing the identity of the persons to whom they belong) to unnoticed output bias that caused racist discriminatory issues~\cite{whittaker2018ai}. For~instance, data-based models for vehicular perception, obstacle detection and avoidance must be also endowed with ethics and legal design factors to make the overall decision not just drifted by the data themselves. Another clear domain where FATE can be crucial is modeling with crowd-sourced Big Data, where~aspects like privacy preservation~\cite{victor2016privacy} and bias avoidance~\cite{rashidi2017exploring} are arguably more critical~\cite{boyd2012critical,chen2017traffic}. The~construction of the data-based modeling workflow must (i)~ensure that protected features remain as such once the workflow has been built, without any chance for reverse engineering (via e.g.,~XAI techniques~\cite{arrieta2020explainable}) that could compromise the differential privacy of data; and (ii)~that learning algorithms along the workflow counteract hidden bias in data that could eventually lead to discriminatory decisions (due to skewed samples, tainted annotation, limited data sizes or imbalanced data). From our perspective, these are among the most concerning challenges in the exploitation of Big Data in ITS, and~the main source of motivation for a number of recent studies in areas related to data-driven transportation systems such as pedestrian detection~\cite{wilson2019predictive}, autonomous vehicles~\cite{lim2019algorithmic,bigman2020life} or urban computing~\cite{fu2019batman}. Bias-related issues can be identified by a proper analysis of the decisions made by the workflow, which in turn requires models to be accountable and transparent enough to thoroughly characterize their sensitivity to bias, and~how inputs and outputs (decisions) correlate in regards to protected features. It is also remarkable to note that several proposals have been made to quantify fairness in machine learning pipelines, yielding useful metrics that account for the parity of models when processing groups of inputs~\cite{leben2020normative,verma2018fairness}. Without these aspects being considered jointly with performance measures, data-based ITS developments in years to come are at the risk of being restricted to the academia playground~\cite{zook2017ten}.

\subsection{Other AI Research Areas Connected to Actionability}

The above areas have been highlighted as the main propellers for model actionability in ITS systems. However, it is worthwhile to mention other research areas from the AI realm that can also help completing the chain from data to actions:
\begin{itemize}[leftmargin=*]
	\item Few-shot learning~\cite{fei2006one}, which aims at overcoming the lack of reliably annotated data and the practical difficulty of performing annotation in certain application scenarios. For~instance, accident prevention models cannot be enriched with positive samples unless a fatality occurs and the data captured in place is fed {into} the model. Few~shots learning and related subareas (zero-shot, one-shot) deriving solutions that can automatically learn from very small amounts of training data, incorporating mechanisms (e.g.,~generative models, regularization techniques, guided simulation) to prevent the overall model from overfitting~\cite{1904.05046}. In~regards to actionability, this~family of learning techniques can be helpful to make data-based ITS models deployable in situations lacking data supervision, specially when such a data annotation cannot be guaranteed to be achievable over time.
	\item Imbalanced and cost-sensitive learning~\cite{krawczyk2016learning,branco2016survey}, which link to the need for avoiding model bias, not only to ensure the generalization of its output, but also to reduce the likeliness of the workflow to cause discriminatory issues as the ones exemplified above. The~history of these AI areas in the ITS community has been going for years now~\cite{zhang2011data}. However, we~here emphasize the crucial role of these techniques beyond performance boosting: the techniques originally aimed to counteract the effects of class imbalance in the output of data-based models could be also leveraged to reflect legal impositions that not necessarily relate to the model's performance nor can they be inferred easily for the attributes within the data themselves. The~lack of compliance of the model with fairness and ethics standards does not necessarily render a performance degradation observed at its output, nor can it be inferred easily from the available data.
	\item Hybrid models encompassing linguistic rules and data-based learning techniques, capable of supporting the transition from the traditional way of doing to the new data-based modeling era in the management of ITS systems. We~foresee that the community will witness a renaissance of data mining methods incorporating methods such as fuzzy logic not only to implement human knowledge to decision workflows, but also to explain and describe the internal structure of learned models, as it is currently under investigation in many contributions under the XAI umbrella~\cite{fernandez2019evolutionary,mencar2018paving}.
	\item New prescriptive data-based techniques such as Deep Reinforcement Learning~\cite{1701.07274} and Algorithmic Game Theory~\cite{nisan2007algorithmic} will also grasp interest in the near future for their close connection to actionable data science. The~interaction of data-based workflows with humans will require techniques capable of learning actions from experience, and~eventually orchestrating the interaction and negotiation among users when their actions are governed by interrelated yet conflicting objectives. In~fact such new prescriptive elements are progressively entering the literature in certain ITS applications that target machine autonomy (e.g.,~autonomous vehicle~\cite{sallab2017deep,ruch2019value} or automated signaling~\cite{mannion2016experimental}), but it is our vision that they will gain momentum in many other ITS setups.
	\item Privacy-preserving Data Mining~\cite{aldeen2015comprehensive,agrawal2000privacy}, which has garnered a great interest in the last year with major breakthroughs reported in the intersection between machine learning, cryptography, homomorphic encryption, secure enclaves and blockchains~\cite{mendes2017privacy}. The~use of personal data and the stringent pressure placed by governments and agencies on differential privacy preservation has spurred a flurry of research to prevent models from revealing sensitive data from their training instances~\cite{ding2019survey, victor2016privacy}. Within the ITS domain, it is possible to find many areas in which privacy preservation has recently been a subject of intense research: from origin-destination flow estimation~\cite{zhou2013privacy} to route planners~\cite{florian2014privacy, rabieh2015privacy}, or pattern mining~\cite{kim2008privacy}, a~glance at recent literature reveals the momentum this topic has acquired lately. In~any of these examples data are available as a result of the sensing pervasiveness (specially in the case of VANETs) and the capture of user data. While previous works explored how to used these data in a proper way with respect to privacy matters, it is straightforward to think that the natural evolution of this research line arrives at how protected data is preserved through the modeling workflow.
	\item Furthermore, the~proven vulnerability of data-based models against adversarial attacks has also motivated the community to lay the foundations of an entirely new research area---Adversarial Machine Learning---, 
	committed to the design of robust models against attacks crafted to confuse their outputs~\cite{huang2011adversarial,1312.6199}. Interestingly, one of the most widely exemplified scenarios in this research area relates to ITS: automated traffic signal classification models were proven to be vulnerable to adversarial attacks by placing a simple, intelligently designed sticker on the traffic sign itself~\cite{akhtar2018threat}. Likewise, the~rationale behind Federated Learning (discussed in Section~\ref{sec:tfada}) also spans beyond the efficient distribution of locally captured knowledge among models: since no raw data instances are involved in the information transfer, privacy of local data is consequently preserved. In~short: security also matters in actionable data-based pipelines.
	\item Finally, the~ever-growing scales of ITS scenarios demand more research invested in scaling up learning algorithms in a computationally efficient manner~\cite{nguyen2019machine}. Automated traffic, smart cities, mobility as a service constitute ITS scenarios where a plethora of information sources interact with each other. Definitely more efforts must be invested in aggregation strategies for data-based models learned from different interrelated data ecosystems, either in a distributed fashion (e.g.,~federated learning) or in a centralized system (correspondingly, Map-Reduce implementations of data-based models, cloud-based architectures, etc). Computational aspects of large-scale implementations should be also under study due to their implications in terms of actionability, such~as the latency of the system when prescribing decisions from data. This latter aspect can be a key for real-time ITS applications for which the gap from data to actions must be shortened to its minimum.
\end{itemize}

\section{Concluding Notes and Outlook}\label{conclusion}

This work has built upon the overabundance of contributions within the ITS community dealing with performance-based comparisons among data-based models. Our claim is that, as in any other domain of application, data-based modeling should bridge the gap between data and actions, providing further value to the ITS application at hand than superior model performance statistics. It is our firm belief that the research community should embrace actionability as the primary design motto, with negligible performance improvements being left behind in favor of relevant aspects such as adaptability, usability, resiliency, scalability or efficiency.

To provide a solid rationale for our postulations, we~have first presented a reference model for actionable data-based workflows, placing emphasis on the different phases that should be undertaken to translate data into actions of added value for the decision maker. Adaptation has been highlighted as a necessary albeit often neglected processing step in data-based modeling, which allows models to be effective when deployed on dynamic ITS environments with time-varying data sources. Next, our study has listed the main functional requirements that models along the reference model should meet to guarantee their actionability, followed by an overview of incipient research areas in Data Science and Artificial Intelligence that should progressively enter the ITS arena. Indeed, advances in XAI, Online Learning, Gray-box Modeling and Transfer Learning are currently investigated mostly from an application-agnostic perspective. Their undoubted connection to actionability makes them the core of a promising future for data-based modeling in ITS systems, processes and applications. 

{ Other research areas related to Artificial Intelligence beyond those covered in our reflections will surely spawn further opportunities for actionability in ITS, provided that they fully embrace their ultimate goal: to effectively support decision making. Among them, the~use of Automated Machine Learning (AutoML~\cite{hutter2019automated}) for tuning data-based models should not only optimize performance-based metrics (e.g.,~finding a model that attains maximum accuracy for image segmentation in vehicular perception cameras), but also comply with other objectives and constraints that closely link to actionability (e.g., robustness against adversarial attacks, or a lower epistemic uncertainty of the model induced in its output). Unless all such actionability constraints are regarded as design objectives and accounted for as such in the automated discovery of new data-based pipelines, any incursion of AutoML in ITS will be of no practical value. For~this to occur, it is our belief that the confluence of multiple functional and non-functional requirements in this automated design process will pave the way towards the massive adoption of multi-objective optimization algorithms as a massive framework to infer and analyze all trade-offs existing among the design objectives.}

Data-based modeling has brought a deep transformation to ITS. A~vast amount of research works in the field are produced by data-based modeling specialists attracted by the profusion of available data, and~with limited knowledge of transportation. Data-based models are getting progressively more complex, increasing the gap between research and practice. This situation calls for a change of paradigm, to a one in which actionability requirements of models is desired by researchers, and~practitioners are aware of the technologies available to provide it. Model actionability is a great whole that can act as an incentive to perform smaller steps towards its realization. It is probably unthinkable to develop, in~a research environment, a~data-based model that meets all proposed requirements. However, addressing some of the postulated requirements while developing a competing data-based ITS model will make it closer to actionability. There is, therefore, a~long road to be travelled in ITS model actionability, with interesting avenues around the thorough understanding of models, and~the adoption of emerging AI technologies to endow data-based workflows with the requirements needed to make them actionable in practice. As exposed in our study, there is a germinal interest in these research topics. Nevertheless, we~foresee vast opportunities for future work when model actionability is set as a design priority.

On a closing note, we~advocate for a new dawn of Data Science in the ITS domain, where advances in modeling performance concurrently emerge along with histories and reports about how such models have helped decision making in practical scenarios. Data mining has limited merit without actions prescribed from its outputs, always in compliance and close match with the specificities of its context.

\vspace{6pt} 





\section*{Acknowledgements} 

This work was supported in part by the Basque Government for its funding support through the EMAITEK program (3KIA, ref. KK-2020/00049). It has also received funding support from the Consolidated Research Group MATHMODE (IT1294-19) granted by the Department of Education of the Basque Government.







\end{document}